\documentclass[10pt, a4paper]{article}

\usepackage{lrec-coling2024}

\usepackage{times}
\usepackage{latexsym}
\usepackage{graphicx}
\usepackage{amsmath}
\usepackage{cleveref}
\usepackage{float}
\usepackage{booktabs}
\usepackage{arydshln}
\usepackage{hyperref}

\usepackage{xcolor}

\usepackage[T1]{fontenc}

\usepackage[utf8]{inputenc}

\usepackage{microtype}

\usepackage{inconsolata}

\newcommand{\model}{\textsc{HyperTTS}}

\title{\model{}: Parameter Efficient Adaptation in Text to Speech using Hypernetworks}

\name{Yingting Li$^{1}$\(^*\)\thanks{\(^*\) Equal Contribution.}, Rishabh Bhardwaj$^{2}$\(^*\), Ambuj Mehrish$^{2}$\(^*\), Bo Cheng$^{1}$, Soujanya Poria$^{2}$}

\address{$^{1}$Beijing University of Posts and Telecommunications, China \\
        $^{2}$Singapore University of Technology and Design, Singapore \\
         \texttt{cindyyting@bupt.edu.cn, rishabh\_bhardwaj@mymail.sutd.edu.sg, ambuj\_mehrish@sutd.edu.sg} \\
         \texttt{chengbo@bupt.edu.cn, sporia@sutd.edu.sg} \\
         }

\abstract{
Neural speech synthesis, or text-to-speech (TTS), aims to transform a signal from the text domain to the speech domain. While developing TTS architectures that train and test on the same set of speakers has seen significant improvements, out-of-domain speaker performance still faces enormous limitations. Domain adaptation on a new set of speakers can be achieved by fine-tuning the whole model for each new domain, thus making it parameter-inefficient. This problem can be solved by Adapters that provide a parameter-efficient alternative to domain adaptation. Although famous in NLP, speech synthesis has not seen much improvement from Adapters. In this work, we present \textbf{\model{}}, which comprises a small learnable network, ``hypernetwork", that generates parameters of the Adapter blocks, allowing us to condition Adapters on speaker representations and making them dynamic. Extensive evaluations of two domain adaptation settings demonstrate its effectiveness in achieving state-of-the-art performance in the parameter-efficient regime. We also compare different variants of \model{}, comparing them with baselines in different studies. Promising results on the dynamic adaptation of adapter parameters using hypernetworks open up new avenues for domain-generic multi-speaker TTS systems. The audio samples and code are available at \url{https://github.com/declare-lab/HyperTTS}.
\\ 
\newline  }

\begin{document}

\maketitleabstract

\section{Introduction}
Neural text-to-speech (TTS) synthesis has transformed our interactions with digital content by converting text into natural-sounding speech. Current TTS systems are often limited to predefined speaker styles or specific sets of speaker IDs \cite{ren2019fastspeech}, reducing their utility in multi-speaker environments with unseen speakers. To make TTS scalable and economical, parameter-efficient adaptation of such systems to new speakers is an important, but highly challenging problem \cite{li2023evaluating}.


Zero-shot and few-shot speaker adaptation techniques \cite{shen2023naturalspeech,li2023styletts,DBLP:conf/interspeech/CasanovaSGMOCSA21,cooper2020zero,casanova2022yourtts,shen2023naturalspeech} have gained prominence in the domain of TTS, aiming at accommodating new speakers and styles with limited speaker-specific data. While these methods excel in scenarios with constrained data, it's important to note that when sufficient data is available, fine-tuning the model offers distinct advantages. Fine-tuning allows for highly personalized and tailored speech synthesis, precise control over the alignment of synthesized speech with the speaker's characteristics, and the production of higher-quality, more natural-sounding speech.
\begin{figure}
    \centering
    \includegraphics[width = 0.8\columnwidth]{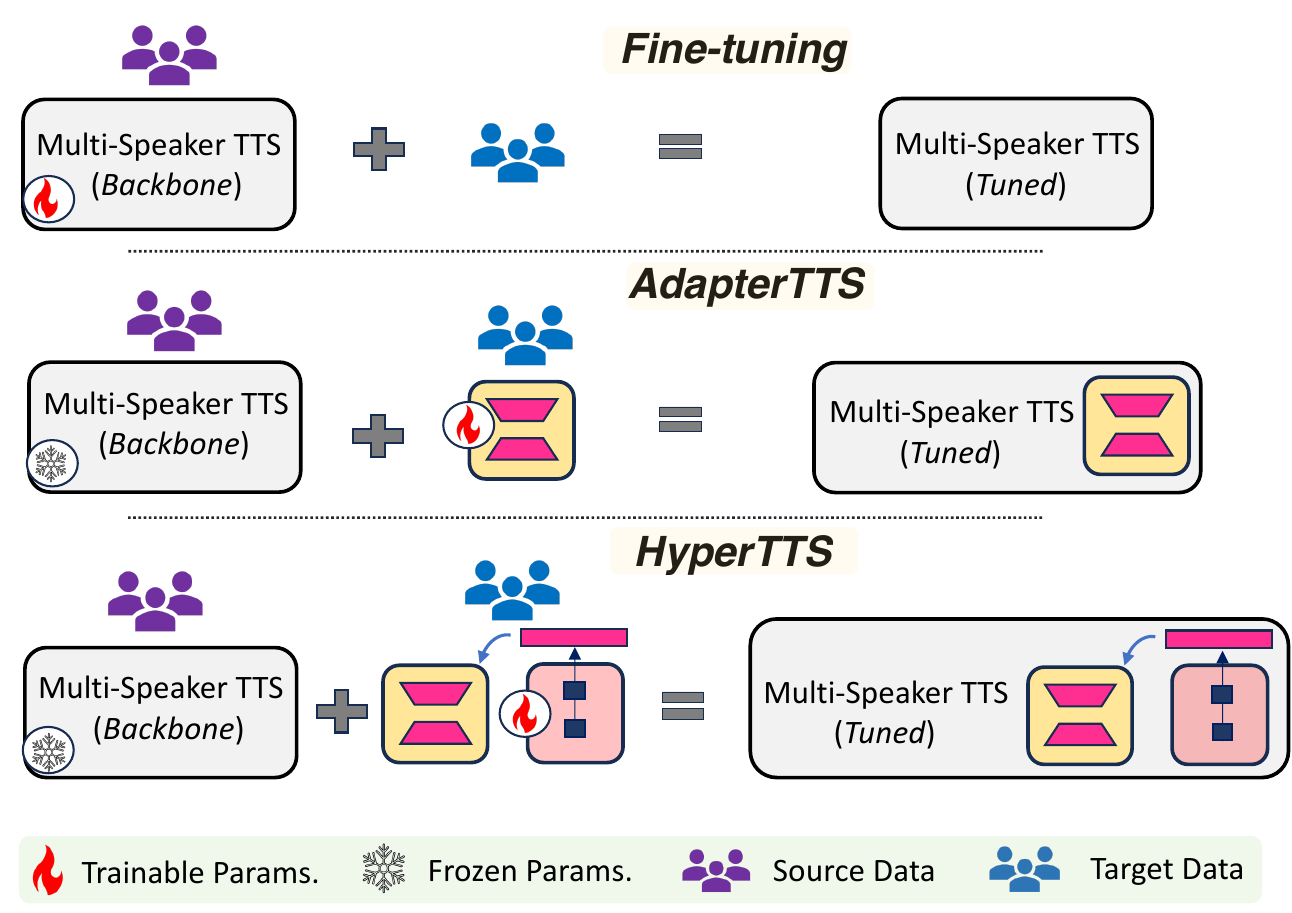}
    \caption{\footnotesize Comparison of our approach against baselines: \textit{Fine-tuning} tunes the backbone model parameters on the adaptation dataset. \textit{AdapterTTS} inserts learnable modules into the backbone. \textit{HyperTTS} (ours) converts the static adapter modules to dynamic by speaker-conditional sampling using a (learnable) hypernetwork. Both AdapterTTS and HyperTTS keep the backbone model parameters frozen and thus parameter efficient.}
    \label{fig:introduction}
\end{figure}

In this paper, we assume sufficient availability of data from the adaptation domain. When adapting a multi-speaker TTS model (\textbf{backbone model}) to a target domain, the traditional approach involves complete fine-tuning of the entire backbone (\Cref{fig:introduction}-\textit{Fine-tuning}). However, this approach is resource-intensive, requiring separate copies of model parameters for each new target domain. To make the adaptation scalable, recent research has introduced parameter-efficient domain adaptation methods using Adapters, as seen in NLP \cite{houlsby2019parameter} and speech \cite{li2023evaluating}. Adapters incorporate small blocks of learnable dense layers into each block of the backbone model, with the aim of learning additional parameters while keeping the main model parameters fixed (\Cref{fig:introduction}-\textit{AdapterTTS}). Despite the advantages demonstrated by adapters in various NLP tasks, their direct application in adapting a TTS backbone to a target domain has shown limited improvements \cite{li2023evaluating}

Since learning a generic TTS system that works well across different speaker styles is a more difficult problem than learning one network per speaker \cite{ren2019fastspeech, DBLP:conf/iclr/0006H0QZZL21}, we hypothesize the same is the case with adapters. Forcing a static set of adapter parameters to perform well across multiple speakers of the adaptation domain can be challenging and potentially infeasible due to under-parameterization \cite{mehrish2023adaptermix,DBLP:conf/interspeech/BiadsyCZRRM22}. 


In this paper, we present \textbf{\model{}}, a pioneering approach for the parameter-efficient adaptation of TTS models to new speakers. This method conditions adapters on speaker embeddings, expanding the learnable parameter space through a "hypernetwork". The main highlights of HyperTTS are:
\begin{enumerate}
    \item \textit{Dynamic Adapters}: Instead of keeping the adapters static, for each speaker in the adaptation domain, \model{} learns speaker-adaptative adapters. Adapter conditioning on speaker representations is observed to unlock adapter capabilities and make them performant which was a challenge with static adapters \cite{li2023evaluating}.
    \item \textit{Parameter Sampling}: A large set of speakers makes it infeasible to keep the space of adapter parameters discrete. To facilitate this, we employ parameter sampling from a continuous distribution defined by a learnable hypernetwork.
    
    \item \textit{Parameter Efficiency}: Compared to parameter-expensive fine-tuning, it achieves competitive results with less than $1$\% of the backbone parameters, making it highly practical and resource-friendly for scalable applications. 
\end{enumerate}


We perform a comprehensive set of experiments to showcase \model{}'s effectiveness (see \Cref{fig:introduction}) compared to traditional methods like static bottleneck adapters (AdapterTTS) and full model fine-tuning (TTS-FT). Our experiments cover datasets from diverse environmental conditions, such as LibriTTS and VCTK, representing various accents from different regions. Results highlight \model{}'s parameter-efficient performance advantages over the baselines across both objective and subjective metrics. Notably, \model{} can even surpass fine-tuning in performance with only a 20\% increase in parameters (\Cref{tab:hypernetwork_ablations_dimension_all}-\model{}$_{e/v/d}$). A key strength of \model{} lies in its remarkable parameter efficiency: it achieves results within 1 point of fine-tuning while using less than 1\% of the parameter count in the backbone. This practical and resource-friendly approach enables real-world applications.


\begin{figure*}[ht]
    \centering
    \includegraphics[width=0.8\linewidth]{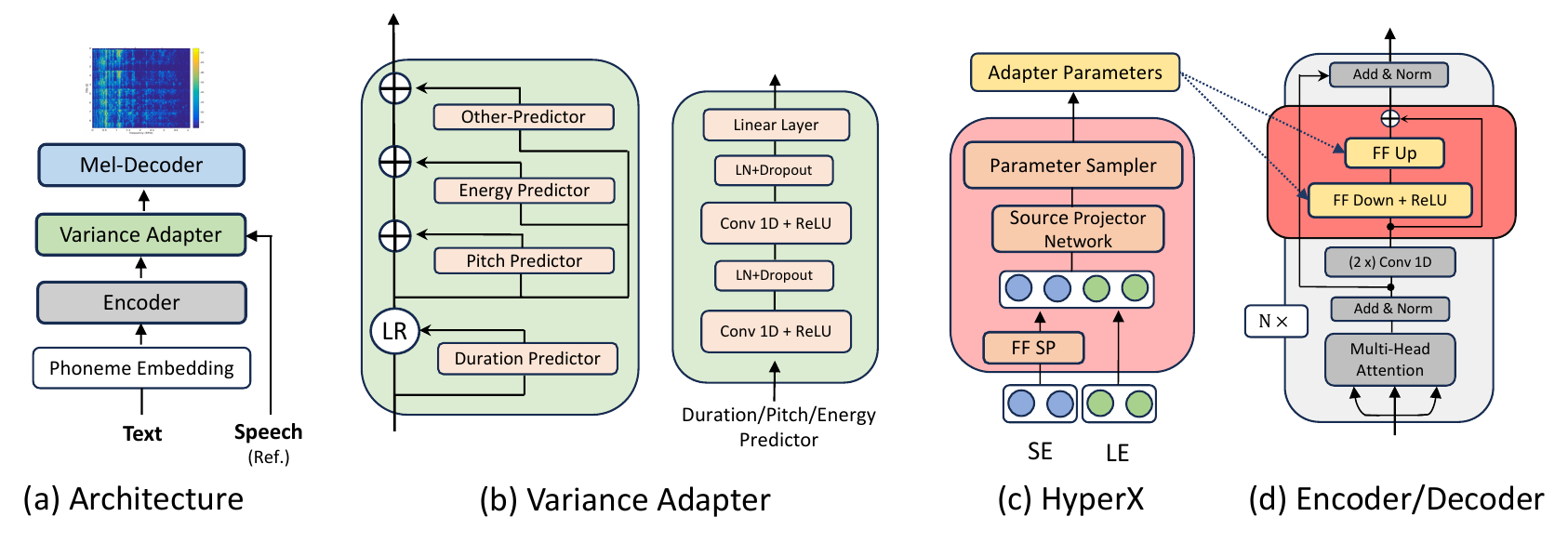}
    \caption{\footnotesize An overview of the \model{}. SE and LE denote speaker embedding and layer embedding.}
    \label{fig:architecture}
\end{figure*}


\section{Related Works}
\paragraph{Text-to-speech models.}

The rise of deep learning has transformed TTS technology, with neural network-based architectures like Tacotron \cite{Tacotron,Tacotron2}, FastSpeech2 \cite{DBLP:conf/iclr/0006H0QZZL21}, and Transformer-TTS \cite{li2019neural} leading the way. These models represent significant progress in TTS, leveraging deep learning techniques. Autoregressive TTS models \cite{Tacotron, Flowtron, DBLP:conf/iclr/0006H0QZZL21, ren2019fastspeech, kim2020glowtts, 9413889}, while effective, face limitations in maintaining alignment in long utterances and exhibit slower training and inference speeds with longer sequences. In contrast, non-autoregressive (parallel) models separate phoneme duration estimation from decoding, reducing latency and enhancing training efficiency. These models typically rely on external aligners or pre-trained autoregressive models for phoneme duration. To achieve training efficiency and support end-to-end TTS, this paper focuses on a non-autoregressive TTS model with an alignment framework based on the RAD-TTS \cite{rad_tts}  alignment learning objective, as proposed by \citet{badlani2022one}.
Recently, several speech models have been compared to GPT in natural language processing, with a focus on in-context learning for speech. Notably, VALL-E \cite{wang2023neural} and SPEAR-TTS \cite{kharitonov2023speak} leverage emerging codecs to learn discrete speech tokens and employ a vocoder-like decodec to convert these tokens into waveforms. Meanwhile, Voicebox, inspired by flow-matching and aligned with the Fastspeech framework, utilizes continuous features like Mel spectrogram and HiFi-GAN.

\paragraph{Speaker Adaptation in TTS.}
Speaker adaptation is a crucial aspect of TTS systems, aiming to personalize the synthesized speech by modifying the voice characteristics to match those of a specific target speaker. Over the years, various techniques and approaches have been proposed to address the challenges associated with speaker adaptation in TTS \cite{jia2018transfer,chenadaspeech,min2021meta,Hsieh2022AdapterBasedEO,gabrys2022voice}. Furthermore, several studies have focused on exploring parameter-efficient methods for adapting TTS to new sets of speakers, addressing the need for effective adaptation in diverse speaker scenarios. These approaches aim to accommodate a wide range of linguistic variations \cite{pamisetty2023lightweight,do2022text}, including diverse accents \cite{yang2023parameter}, speakers \cite{luo2021lightspeech,miao2021efficienttts,mehrish2023adaptermix}, and low-resource scenarios introduced by the target domain \cite{azizah2022transfer,mehrish2023adaptermix,lux2022language}, while maintaining the number of trainable parameters. \model{} primarily focuses on contributing in the line of parameter-efficient domain adaptation of the backbone TTS model to a target set of speakers. 
 
 \textbf{Dynamic Parameters.} Parameter generation, although not popular in speech, has been used in various forms in other domains, such as \citet{Klein2015ADC, Riegler2015ConditionedRM} in NLP and \citet{Ha2017HyperNetworks} in computer vision. Specific to adapters, \citet{bhardwaj-etal-2022-vector, Chen2020MultiSpeechMT} make prompt tokens dynamic by conditioning their values on input text using a parameter prompt generator network, \cite{ustun2022hyper, mahabadi2021parameter} used hypernetworks for generating adapter down and up-projection weights. Shared hypernetworks obviate the need to maintain a separate set of parameters for each task (or new setting) and generate weights for each block of the backbone network \cite{mahabadi2021parameter}. To the best of our knowledge, this is the first work that studies the utility of a parameter generator in the domain of speech \cite{mehrish2023review}.

\section{Methodology}
The TTS backbone architecture comprises a text (phoneme) encoder, Variance Adapter, and Mel-Decoder. We pre-train the backbone on LibriTTS, a large multi-speaker dataset for which we provide details in \Cref{sec:backbone}.

\subsection{Encoder}
Given a phoneme sequence $(p_1, \ldots, p_n)$ obtained from text, we first map it to vector embedding mixed with sinusoidal positional encoding. The Encoder is composed of four Feed-Forward Transformer (FFT) blocks, with each block comprised of two multi-head attention modules and two 1D-convolutions. Shown in \Cref{fig:architecture}-(d), we adopt Encoder's FFT from FastSpeech \cite{ren2019fastspeech} which adapted the text-specific transformer block from \citet{vaswani2017attention} for text-to-speech. Each head performs a self-attention to contextualize each phoneme with relevant global phoneme sequence information. Followed by contextualization with multiple heads, the encoder uses two 1D convolutions over the phoneme sequence to capture local phoneme information. This is because adjacent phonemes and mel-spectrogram features are more closely related in speech, thus, kernels are intuitively an effective alternative\footnote{We do not claim convolutions are better replacements of dense layers. The model characteristics can change with large parameter speech models, thus leaving it for future research.}. The output of the first convolution is ReLU activated.

\subsection{Variance Adapter}
Variance Adapter (VA) (\Cref{fig:architecture}-(b)) transforms the phoneme embeddings of length $n$ into mel-spectrogram embeddings of length $m$ where $m$ is typically larger than $n$. Thus, each phoneme at the input tends to map to one or multiple mel-frames. To solve this length mismatch, VA comprises a duration predictor that predicts how many mel-frames are required for each phoneme. Following this, VA also dedicates itself to predicting pitch and energy for each element of the length-regulated phoneme sequence. Each predictor is associated with a corresponding loss function.
\paragraph{Duration Predictor.} 
Let $\mathcal{H}=(h_1, \ldots, h_n)$ denote the phoneme sequence input of VA. Duration predictor takes this as the input and predicts $\mathcal{D}=(d_1, \ldots, d_n)$ where $d_i$ is a positive integer that tells how many mel-frames the phoneme will be mapped to. Before feeding to Pitch and Energy, the phoneme sequence is expanded using predicted phoneme durations $\mathcal{D}$. The length-regulated phoneme sequence will appear $(h_1 .. [d_1\text{-times}] .. h_1, \ldots, h_n .. [d_n\text{-times}] .. h_n])$.

Duration predictor warrants ground truth durations, i.e., alignment between phoneme sequences and mel-frames. Following \citet{badlani2022one}, we train a phoneme-to-mel unsupervised alignment predictor end-to-end with the backbone TTS architecture, obviating the need for external aligners such as MFA \cite{MFA} and non-autoregressive aligner models \cite{peng2020non, ren2019fastspeech, 9413889}. Similar to \citet{rad_tts}, we compute the soft alignment distribution based on the learned pairwise affinity between all text tokens and mel-frames. Given a soft alignment map, the Viterbi algorithm is used to find the monotonic path with the highest likelihood in order to convert soft alignments to hard alignments. We relegate specifics of the duration predictor to \citet{badlani2022one, rad_tts}.
\paragraph{Pitch Predictor.}
We use continuous wavelet transform (CWT) for pitch contour prediction, following the approach in \citet{DBLP:conf/iclr/0006H0QZZL21}. During training, the pitch predictor targets a pitch spectrogram obtained from the continuous pitch series decomposition. During inference, the pitch predictor's spectrogram output is converted back to pitch contours using inverse CWT (iCWT). This pitch predictor comprises a two-layer 1D-convolutional network with ReLU activation, followed by layer normalization and dropout. The pitch spectrogram is obtained via a linear layer. To handle discontinuous pitch contours, we employ linear interpolation for unvoiced frames, apply a logarithmic transformation, and normalize to have zero mean and unit variance. Mean and variance predictions are extracted from the output of the 1D-convolutional layer across the time dimension, and values are obtained from separate linear layers. The pitch predictor's parameters are learned by minimizing the mean squared error (MSE) between the spectrogram, mean, and variance values of the ground truth and predictions

\paragraph{Energy Predictor.}
The energy predictor estimates the original energy values for each Short-Time Fourier Transform (STFT) frame, for which it computes the L2-norm of the frame's amplitude. To optimize efficiency, the calculated energy is quantized into 256 evenly distributed values. These quantized values are then encoded into an energy embedding and added to the expanded hidden sequence, similar to the pitch information.
\subsection{Mel-Decoder and Postnet.}
The decoder converts the variance adaptor's hidden sequence into a mel-spectrogram. It shares the same architecture as the encoder but with six FFT blocks. To improve mel-spectrogram quality, a Postnet is used at the mel-decoder's output, reducing artifacts and distortions in speech.

\subsection{Hypernetwork} \label{sec:hypernetwork}
The adapter performs the following operation:
\begin{equation}
    h = h + ReLU(hW_{d})W_{u}
\end{equation}
where $h$ is the hidden representation in the TTS network, $W_{d}$  and $W_{u}$  are down-projection ($d_{h}\xrightarrow{}d_{r}$) and up-projection matrices ($d_{r}\xrightarrow{}d_{h}$) with $d_r$ as bottleneck dimension. The projection matrices are attached to each layer of the encoder, decoder, and variance adapter after the convolutional layers (shown in Figure 2-d). Although adapters have shown significant advantages in NLP and several speech tasks \cite{houlsby2019parameter,li2023evaluating}, they are observed to be less useful in performing TTS adaptation \cite{li2023evaluating}. We posit that forcing a static set of adapter parameters to perform well across multiple speakers of the adaptation domain can be challenging.

\model{} is aimed to make adapters significantly more effective by conditioning them on speaker embeddings, thus enhancing the (effective) learnable parameter space of adapters. For this purpose, it employs a hypernetwork. A hypernetwork is typically a small neural network that generates weights for a larger main network performing the usual learning task (here it is text-to-speech). We propose to enhance the effectiveness of adapters by learning to adapt their parameters with the change in speaker:
\begin{align}
    W_{d} &= f_{d}(v_{s},v_{l})\\
    W_{u} &= f_{u}(v_{s},v_{l})
\end{align}
Here $v_{s}$ and $v_{l}$ denote speaker embedding of dimension $d_{1}$ and layer embedding of dimension $d_{l}$, respectively. $f_{d}$ and $f_{u}$  denote parameter generators which are part of the hypernetwork and are learnable modules. Notably, to keep the network small in size, we leverage a shared hypernetwork to generate parameters for adapters in every layer of a given module of the TTS backbone. We only train the hypernetwork while keeping the TTS backbone frozen. Since the hypernetwork is significantly smaller than the backbone, we refer to the adaption as parameter efficient. In this work, the hypernetwork is smaller than 1\% of the TTS backbone size (in the number of parameters).

\paragraph{Implementation:} With reference to Figure 2-c, we project $d_{1}$-dimensional speaker embedding (SE) onto a $d_{2}$-dimensional space, where $d_{2} < d_{1}$, using a speaker projector (SP) which is a feedforward network with bias. To this, we concatenate a $d_{l}$-dimensional layer embedding (LE). Layer embeddings are a learnable look-up table that maps a layer-id to a vector. The concatenated vector is passed through a source projector network that maps a ($d_{2} + d_{l}$)-dimensional vector into a $d_{s}$-dimensional space. For adapter down and up projection, we sample weights from the hypernetwork through the source projector network using dedicated dense (Parameter Sampler) layers.
\section{Experiments}

In this section, we elaborate on the comparable baselines, target domain multi-speaker datasets, TTS backbone model configurations, and evaluation metrics used.
\subsection{Baseline models}
We assess the quality and similarity of speech samples generated by \model{} by comparing them to other methods. These methods include:
\paragraph{TTS-0} TTS-0 represents the zero-shot performance of the TTS model, wherein it is pre-trained on source data (LTS) and subsequently evaluated on target data without any fine-tuning, which is kept as a baseline to define lower-bound on the performance of the parameter-efficient training (\model{} and AdapterTTS).
\paragraph{Reference and Reference (Voc.)} Reference refers to ground truth speech. To obtain Reference (voc.) we transform the reference speech into mel-spectrograms and subsequently reconstruct the speech signals using HiFi-GAN \cite{kong2020hifi}.

\paragraph{TTS-FT (full fine-tuning)} TTS-FT denotes the model obtained after fine-tuning all parameters of the backbone model on the target dataset, which is kept as a baseline to define upper-bound on the performance of the parameter-efficient training (HyperTTS and AdapterTTS).
\paragraph{AdapterTTS} It inserts bottleneck adapter modules, a down-projection and up-projection layer, in each layer of the backbone model's encoder. We follow the configuration proposed by \citet{pfeiffer2020adapterfusion}. AdapterTTS learns only adapter parameters, keeping the backbone parameters frozen. We examine various configurations of AdapterTTS: AdapterTTS$_e$, AdapterTTS$_v$, AdapterTTS$_d$ denoting bottleneck adapter block inserted in the encoder, VA, and decoder. The combination of all is represented
by AdapterTTS$_{e/v/d}$.

\paragraph{\model{}} For a comprehensive evaluation, we examine various variants of \model{}: \model{}$_e$, \model{}$_v$, \model{}$_d$ denoting bottleneck adapter block inserted in the encoder, VA, and decoder. The combination of all is represented by \model{}$_{e/v/d}$. Hypernetwork is not shared across the different modules of the architecture, thus, the number of parameters are in order \model{}$_{e/v/d} > $\model{}$_d$ $\approx$ \model{}$_{v}$ $\approx$ \model{}$_{e}$.

\subsection{Datasets}
We primarily base our experiments on English language datasets. For the training and adaptation, we employed three distinct datasets: train-clean-100, the dev-clean and test-clean subset of LibriTTS \cite{DBLP:conf/interspeech/ZenDCZWJCW19}, and VCTK \cite{veaux2013voice}. In the subsequent discussion, we will refer to the train-clean-100 subset as LTS and the (dev-clean, test-clean) subset as LTS2. TTS backbone is pre-trained on the LTS dataset. For adaptation, we consider VCTK and LTS2. Each of the datasets is divided into train and validation subsets. To evaluate the effectiveness of the model, we conducted performance assessments using dedicated validation subsets from both the VCTK and LTS2.
\subsection{Model Configuration} \label{sec:backbone}
 
\paragraph{Backbone Model Pre-training.} 
The encoder consists of four layers, and the decoder comprises six layers, both with a hidden state dimension of 256. In \Cref{fig:architecture}-(a), the variance adapter adds the speaker embedding to the text representation. Speaker embeddings are computed using the speaker verification model trained with the generalized end-to-end (GE2E) loss, incorporating data from LibriSpeech (train-other-500)\cite{panayotov2015librispeech}, Voxceleb1, and Voxceleb2 \cite{nagrani2017voxceleb}. For this work, we set $d_h=256$, $d_r=32$, $d_1=256$, $d_2=64$, $d_l=64$, and $d_s=32$.

In the pre-training of the multi-speaker TTS backbone model on the LTS dataset, speech samples are downsampled to a 16 kHz sampling rate. The model is trained using the Adam optimizer. Unsupervised duration modeling, which includes the pitch predictor, duration predictor, and energy predictor in the variance adapter, begins at 50K steps to enhance model convergence. To ensure stable training, a learning rate warm-up is implemented for the first 4K steps, followed by annealing at steps 300K, 400K, and 500K. After training the backbone model for 600K steps, adaptation is carried out on either the VCTK or LTS2 datasets. Backbone fine-tuning, Hypernetworks in \model{}, and AdapterTTS are adjusted for the next 300K steps. The Hypernetwork is trained for 300K steps using the Adam optimizer with a constant learning rate of 0.0001.


\begin{table*}[!ht]
\centering
\small
\resizebox{0.7\textwidth}{!}{
\begin{tabular}{lccccl}
\toprule
\multicolumn{6}{c}{\textbf{LTS $\rightarrow$ VCTK}}                                                             \\ \midrule
\textbf{Model}    & \textbf{COS} $\uparrow$         & \textbf{FFE} $\downarrow$        & \textbf{WER} $\downarrow$ & \textbf{MCD} $\downarrow$ & \textbf{Params} \\ \midrule
Reference           & $100.000_{(\pm0.000)}$ & $00.00_{(\pm0.00)}$ & $0.2055$            & $-$               & $-$                   \\
Reference (Voc.)    & $95.027_{(\pm0.001)}$  & $22.10_{(\pm0.03)}$ & $0.2074$            & $-$               & $-$                   \\
TTS-0               & $73.794_{(\pm0.004)}$  & $39.19_{(\pm0.02)}$ & $0.2035$            & $5.9232$          & $-$                   \\
TTS-FT              & $80.443_{(\pm0.003)}$  & $34.63_{(\pm0.02)}$ & $0.2027$            & $5.2387$          & 35.7M (100\%)         \\  \midrule
AdapterTTS$_{e}$    & $73.769_{(\pm0.004)}$  & $38.73_{(\pm0.02)}$ & $\textbf{0.2075}$   & $5.9002$          & 66.6K (0.186\%)      \\
AdapterTTS$_{v}$    & $73.131_{(\pm0.004)}$  & $42.87_{(\pm0.02)}$ & $0.2258$            & $6.2733$          & 33.3K (0.095\%)      \\
AdapterTTS$_{d}$    & $76.180_{(\pm0.004)}$  & $39.14_{(\pm0.03)}$ & $0.2101$            & $6.0092$          & 100K (0.280\%)       \\
AdapterTTS$_{e/d}$  & $72.703_{(\pm0.004)}$  & $37.99_{(\pm0.02)}$ & $0.2141$            & $5.8804$          & 166.7K (0.466\%)     \\
AdapterTTS$_{e/v/d}$& $77.298_{(\pm0.006)}$  & $35.53_{(\pm0.03)}$ & $0.2234$            & $\textbf{5.2971}$ & 200K (0.559\%)       \\  \hdashline
HyperTTS$_e$        & $75.432_{(\pm0.004)}$  & $36.07_{(\pm0.02)}$ & $0.2367$            & $5.3930$          & 151.1K (0.423\%)     \\
HyperTTS$_v$        & $73.731_{(\pm0.004)}$  & $38.47_{(\pm0.02)}$ & $0.2367$            & $5.9137$          & 150.9K (0.422\%)     \\
HyperTTS$_d$        & $77.590_{(\pm0.004)}$  & $38.55_{(\pm0.02)}$ & $0.2090$            & $5.9641$          & 151.2K (0.423\%)     \\
HyperTTS$_{e/d}$    & $79.232_{(\pm0.003)}$  & $35.02_{(\pm0.02)}$ & $0.2168$            & $5.3650$          & 302.3K (0.846\%)     \\
HyperTTS$_{e/v/d}$  & $\textbf{79.464}_{(\pm0.003)}$ & $\textbf{34.47}_{(\pm0.02)}$ & $0.2340$   & $5.3293$          & 453.3K (1.269\%)     \\

\bottomrule
\end{tabular}
}
\caption{\footnotesize{Domain adaptation performance on VCTK. TTS-0 denotes the zero-shot performance of the backbone TTS model evaluated on the VCTK test set. TTS-FT is a fine-tuned backbone model on the VCTK train set and evaluated on its test set. Where subscript $m \in \{e,v,d\}$ in \model{}$_{m}$ or AdapterTTS$_{m}$  denotes hypernetwork-adapter or adapter inserted to the encoder, variance adapter, and decoder of the backbone model, respectively. }}
\label{tab:LTS-VCTK}
\end{table*}

\subsection{Evaluation Metrics}
In this section, we discuss various metrics we use to compare our model against the baselines.
\paragraph{Objective Metrics:} We assess timbre and prosody similarity between synthesized and reference audio using two metrics: cosine similarity (COS) and F0 Frame Error (FFE). COS provides insights into speaker similarity by measuring the average cosine similarity between embeddings from the synthesized and ground truth data. FFE focuses on fundamental frequency (F0) information, considering voicing decision error and F0 error metrics. Mel cepstral distortion (MCD) is a widely used objective metric that quantifies perceptual differences between synthesized and original speech. It measures the divergence between the Mel-scale Frequency Cepstral Coefficients (MFCC) of the synthesized speech and the original speech. Additionally, we calculate Word Error Rate (WER) to gauge the intelligibility of the generated speech. In our experiments, we use enterprise-grade speech-to-text (STT) pre-trained silero models \cite{silero_Models} to compute WER.
\paragraph{Subjective Metrics:} In addition to objective evaluations, we conducted a comprehensive series of subjective listening tests with six participants to assess the naturalness of the synthesized audio. These tests offer valuable insights into human perception and subjective preferences. Participants listened to a random selection of 5 sets of 20 sound samples, including four distinct methods (TTS-FT, AdapterTTS$_e$, \model{}$_e$, and \model{}$_d$), along with a reference sample. They rated each sample on a five-point Likert Scale \cite{joshi2015likert}, ranging from 1 (Poor) to 5 (Excellent). To ensure an unbiased evaluation, each row in the test contained samples with identical text content, eliminating potential bias from text variations and allowing for a fair assessment of the samples.

To assess speaker individuality, we conducted an XAB test \cite{mizuno1995voice} to evaluate speaker similarity. The reference speech of the target speaker was denoted as X, while the synthesized speech produced by the AdapterTTS, and $HyperTTS_{d}$ models were presented to listeners as options A and B in random order. A specific comparison was made between AdapterTTS and $HyperTTS_{d}$. A total of 6 listeners with backgrounds in NLP and speech participated in the experiment, during which they were presented with 30 synthesized speech samples (including 15 reference samples).
\begin{table*}[h!]
\centering
\small
\resizebox{0.7\textwidth}{!}{
\begin{tabular}{lccccl}
\toprule
\multicolumn{6}{c}{\textbf{LTS $\rightarrow$ LTS2}}                                                             \\ \midrule
\textbf{Model}     & \textbf{COS} $\uparrow$         & \textbf{FFE} $\downarrow$        & \textbf{WER} $\downarrow$ & \textbf{MCD} $\downarrow$ & \textbf{Params} \\ \midrule

Reference       & $100.000_{(\pm0.000)}$ & $00.00_{(\pm0.00)}$ & $0.2046$           & $-$               & $-$                   \\
Reference (Voc.) & $96.919_{(\pm0.000)}$  & $19.72_{(\pm0.02)}$ & $0.2089$           & $-$               & $-$                   \\
TTS-0           & $78.784_{(\pm0.005)}$  & $43.31_{(\pm0.02)}$ & $0.2129$           & $7.7039$          & $-$                   \\
TTS-FT          & $82.351_{(\pm0.004)}$  & $41.26_{(\pm0.02)}$ & $0.2135$           & $7.5843$          & 35.7M (100\%)         \\  \midrule
AdapterTTS      & $77.989_{(\pm0.005)}$  & $42.28_{(\pm0.02)}$ & $\textbf{0.2143}$  & $7.6581$          & 66.6K (0.186\%)      \\
\model{}$_e$     & $78.302_{(\pm0.006)}$  & $41.38_{(\pm0.02)}$ & $0.2232$           & $\textbf{7.4746}$ & 151.1K (0.423\%)     \\
\model{}$_v$    & $78.853_{(\pm0.005)}$  & $43.19_{(\pm0.02)}$ & $0.2180$           & $7.7055$          & 150.9K (0.422\%)     \\
\model{}$_d$    & $81.021_{(\pm0.005)}$  & $41.77_{(\pm0.02)}$ & $0.2159$           & $7.5942$          & 151.2K (0.423\%)     \\
\model{}$_{e/d}$& $81.360_{(\pm0.005)}$  & $41.95_{(\pm0.02)}$ & $0.2180$           & $7.5865$          & 302.3K (0.846\%)     \\
\model{}$_{e/v/d}$& $\textbf{81.742}_{(\pm0.005)}$ & $\textbf{41.03}_{(\pm0.02)}$ & $0.2337$    & $7.5876$          & 453.3K (1.269\%)     \\

\bottomrule
\end{tabular}
}
\caption{\footnotesize Speaker adaptation on LTS2 is assessed. "TTS-0" represents the base TTS model's zero-shot performance on the LTS2 test set. "TTS-FT" is the fine-tuned base model on the LTS2 training set, evaluated on its test set. The subscript $m \in {e,v,d}$ in \model{}${m}$ or AdapterTTS${m}$ denotes the integration of hypernetwork-adapter or adapter in the encoder, variance adapter, or decoder of the base model, respectively.}
\label{tab:LTS-LTS2}
\end{table*}

\begin{table*}[h!]
\centering
\resizebox{0.7\textwidth}{!}{
\small
\begin{tabular}{@{}lccccl@{}}
\toprule
\textbf{Model}     & \textbf{COS} $\uparrow$         & \textbf{FFE} $\downarrow$        & \textbf{WER} $\downarrow$ & \textbf{MCD} $\downarrow$ & \textbf{Params} \\ \midrule
\model{}$_d$(2)        & $75.89_{(\pm0.0040)}$ & $38.99_{(\pm0.0211)}$ &  $0.2096$   &  $6.0016$   &   50K (0.14\%)
\\  
\model{}$_d$(8)        & $77.59_{(\pm0.0036)}$ & $38.55_{(\pm0.0208)}$ &  $\textbf{0.2090}$   &  $\textbf{5.9641}$   &   151K (0.42\%)
\\
\model{}$_d$(32)       & $79.38_{(\pm0.0033)}$ & $\textbf{38.05}_{(\pm0.0210)}$ &  $0.2152$   &  $6.0024$   &   554K (1.55\%)
\\
\model{}$_d$(128)      & $\textbf{80.26}_{(\pm0.0033)}$ & $38.15_{(\pm0.0210)}$ &  $0.2186$   &  $5.9820$   &   2.17M (6.06\%)
\\ \bottomrule
\end{tabular}
}
\caption{\footnotesize{Varying number of parameters of hypernetwork in decoder with VCTK as the target domain. \model{}$_d$($n$) denotes hypernetwork with $n$-dimensional source projection.}
\label{tab:hypernetwork_ablations_dimension_decoder}}
\end{table*}

\section{Results and Discussions}
In \Cref{tab:LTS-VCTK}, we observe full fine-tuning (TTS-FT) of the backbone TTS model on VCTK (TTS-0) improves the COS score by $\approx$7 points on its test set, and reduce frame error (FFE) of basemodel by $\approx$4.5 points. This forms our baseline in a non-parameter efficient regime. 
First, we carry out extensive experiments on AdapterTTS by inserting it in the encoder, VA, decoder, encoder+decoder, and all three encoder+decoder+VA with approximately 0.2\%, 0.1\%, 0.3\%, 0.5\%, and 0.56\% trainable parameters. We observe that compared with AdapterTTS$_v$ and AdapterTTS$_d$, AdapterTTS$_e$ shows advantages in three of four metrics. Overall, the performance of AdapterTTS across its variants stays significantly lower than fine-tuning performance, making it a less interesting choice for adaptation. While performing TTS adaptation on LTS2 (\Cref{tab:LTS-LTS2}), we draw a similar inference where AdapterTTS$_{e}$ comes out to be hardly advantageous when compared with TTS-FT baselined across the objective metric. Moving further, we experiment on AdapterTTS$_e$ as AdapterTTS$_{e/v/d}$ shows relatively poor performance on WER while the performance of AdapterTTS$_{e/d}$ is lower by 1 point on speaker similarity COS score, which is one of the critical attributes of multi-speaker TTS adaptation.

\begin{table}[t]
\centering
\resizebox{0.7\columnwidth}{!}{
\small
\begin{tabular}{@{}lcccl@{}}
\toprule
\textbf{Model}     & \textbf{MOS} $\uparrow$               &  \textbf{Params} \\ \midrule
Reference        & $4.62_{(\pm0.02)}$     & -                        \\
TTS-FT           & $3.70_{(\pm0.16)}$     & 35.7M (100\%)            \\
AdapterTTS$_e$   & $3.47_{(\pm0.07)}$     & 66.6K (0.186\%)          \\
HyperTTS$_e$     & $3.43_{(\pm0.10)}$     & 151.1K (0.423\%)         \\
HyperTTS$_d$     & $3.64_{(\pm0.06)}$     & 150.9K (0.422\%)         \\
\bottomrule
\end{tabular}
}
\caption{\footnotesize MOS comparison among Reference, TTS-FT, AdapterTTS$_e$, \model{}$_e$ and \model{}$_d$ for samples randomly selected from VCTK validation set.}
\label{tab:subject_eval}
\end{table}
Next, we change the adapter parameters from static learnable to dynamic by sampling them using a hypernetwork, conditioning the sampling process on speaker style and layer-id. In \Cref{tab:LTS-VCTK}, we observe \model{} significantly improves the backbone performance on two out of four metrics while staying in a parameter-efficient regime. \model{}$_d$ i.e., hypernetwork for adapter block in decoder performs best on two out of four metrics across its low parameter setting (<0.5\% added parameters) when comparing against \model{}$_e$ and \model{}$_v$. We posit that adapting \model{}$_e$ is not effective as it is dedicated to encoding phoneme embedding while adapting \model{}$_v$ in VA is observed to be noisy, observed through an increase in MCD score. Trading off with a few parameters ($\approx$1\% added parameters) in \model{}$_{e/d}$ and \model{}$_{e/v/d}$, we can achieve performance close to the full fine-tuning baseline within a margin of $\approx$1\% in COS, FFE, WER, and MCD. Except for WER where decoder adaptation is close to AdapterTTS, \model{} consistently outperforms AdapterTTS and achieves performance closer to the fine-tuning baseline. A similar trend is seen for LTS2 adaptation (\Cref{tab:LTS-LTS2}) where \model{}$_d$ increases the COS score by over 2 points, however, scores did not change significantly on other metrics. We posit that this is potentially due to the undertraining of the network. Similar to the findings of VCTK, we observe while adapting the backbone to LTS2. AdapterTTS worsens in the speaker similarity COS score and MCD although slightly improving on FFE and WER errors. In the class of models with less than 0.5\% added parameters, hypernetwork in decoder \model{}$_{d}$ achieves performance close to the fine-tuning baseline TTS-FT on all the metrics. It is noteworthy that \model{} and adapters while achieving close to the TTS-FT performance, do not inherently suffer from catastrophic forgetting, thus, providing advantages of both regimes. After fine-tuning, TTS-FT performance on the pre-training dataset VCTK becomes worse COS $83.10$ $\rightarrow$ $77.09$   and FFE $39.01$ $\rightarrow$ $39.88$.

\subsection{Subjective Evaluation}
We conducted a subjective evaluation, comparing the naturalness and quality of synthesized speech to a reference sample. Results of the side-by-side subjective test are shown in Table \ref{tab:subject_eval}. Notably, \model{}$_d$ achieved an MOS of 3.6412, surpassing AdapterTTS with an MOS of 3.4727. TTS-FT obtained a MOS of 3.7021 with 35.7 million parameters, while \model{}$_d$ only used 0.422\% of the parameters in TTS-FT, as seen in Table \ref{tab:subject_eval}. These results highlight \model{}$_d$'s superior performance in both subjective evaluation scores and parameter efficiency when compared to AdapterTTS and TTS-FT. We also conducted a paired t-test to assess the statistical significance of our findings. The results indicate a highly significant difference in mean opinion scores between the Reference and all models (p < 0.0001). There is a statistically significant contrast in mean opinion scores between AdapterTTS$e$ ($p\sim0.0415$) and \model{}${e}$ ($p\sim0.0345$) when compared to TTS-FT. However, in the case of \model{}${d}$ ($p\sim0.56$), we lack sufficient grounds to reject the null hypothesis (p < 0.05), suggesting that TTS-FT and \model{}${d}$ likely share identical expected values.

\begin{figure}[h]
    \centering
    \includegraphics[width=\columnwidth]{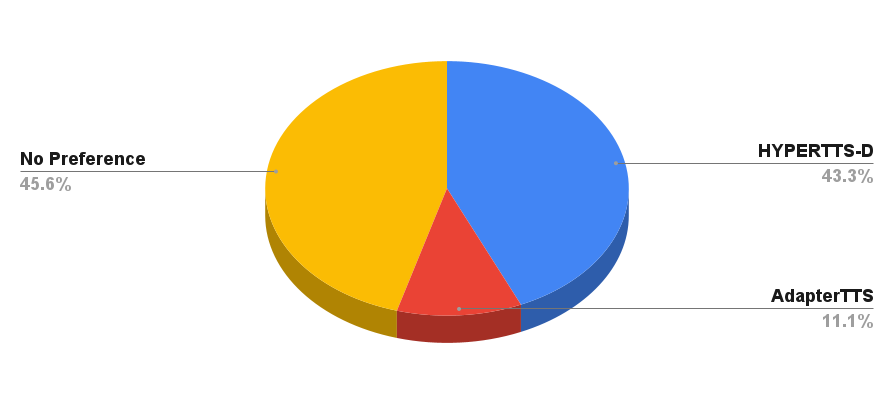}
    \caption{XAB speaker similarity test results between AdapterTTS, and $HyperTTS_{d}$. }
    \label{fig:XAB}
\end{figure}

 The speaker similarity results depicted in \Cref{fig:XAB} indicate a preference for samples generated by  $HyperTTS_{d}$, with a percentage of $43.3$\% favoring it over AdapterTTS.

%

\subsection{Impact of Parameter Efficiency}
The parameter efficiency of hypernetwork and adapters depends on the projection dimension. We investigate this trade-off by varying the size of the source projector network in $3$ variants: \model{}$_e$, \model{}$_d$, and \model{}${e/v/d}$, as shown in \Cref{fig:architecture}-c. The results of these experiments are detailed in~\Cref{tab:hypernetwork_ablations_dimension_decoder,tab:hypernetwork_ablations_dimension_encoder,tab:hypernetwork_ablations_dimension_all}. AdapterTTS, with approximately 66K parameters, has fewer parameters than smaller versions of \model{}. When adapting to VCTK, increasing the down-projection dimension from 32 to 128 results in adapter parameters growing to around 263K. In this case, the COS decreases, while FFE scores increase to 73.58 and 39.46, respectively. This suggests that AdapterTTS does not benefit from parameter scaling in the same way as \model{}.
\subsection{Output of Hypernetwork}  \label{dynamic}
\begin{figure}[t]
    \centering
    \includegraphics[width = 0.8\columnwidth]{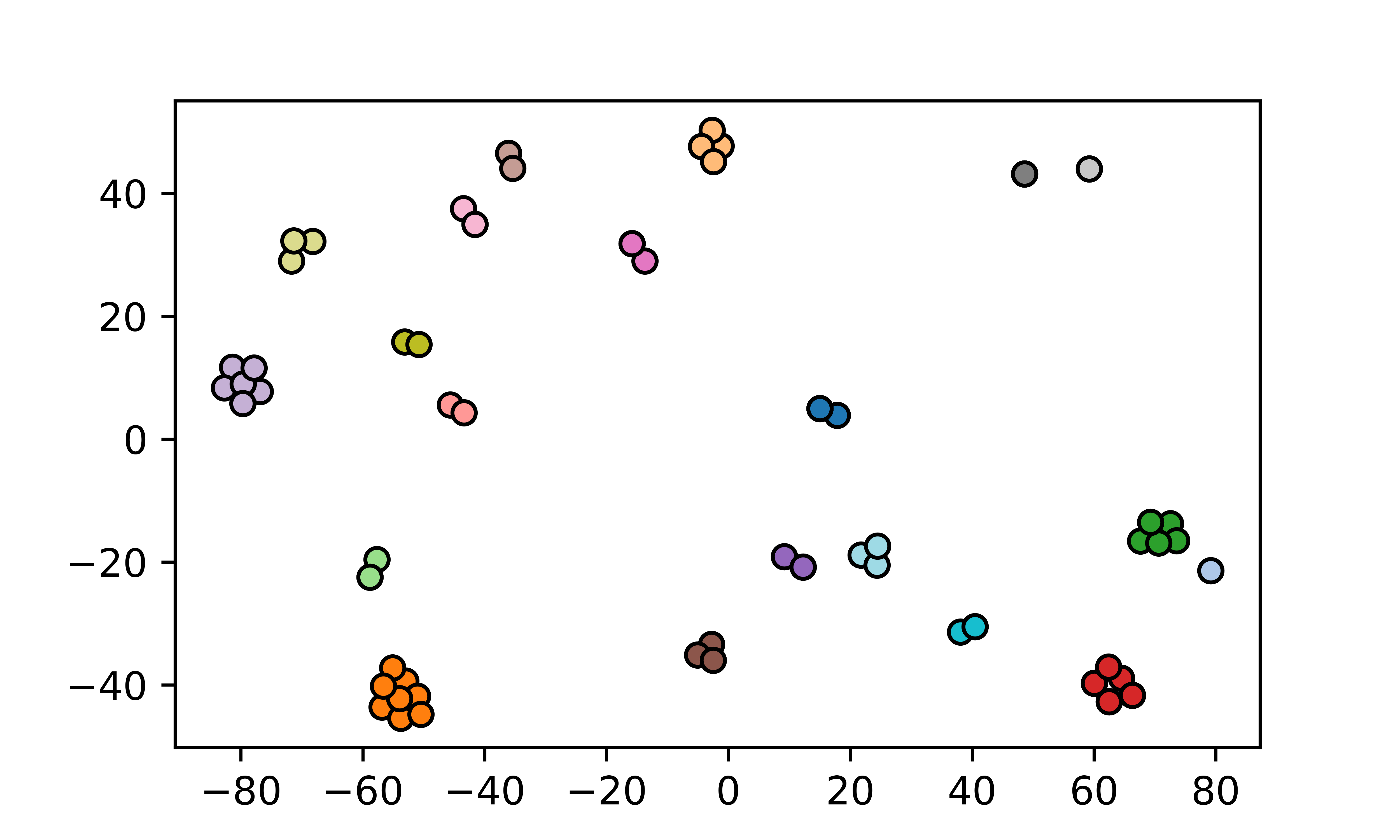}
    \caption{\footnotesize{t-SNE of hypernetwork generated parameters for 20 randomly sampled speakers from VCTK test set. The same color marks represent reference speech from the same speaker.}}
    \label{fig:clustering}
\end{figure}

In \Cref{fig:clustering}, in the continuous space of parameters, we observe that different reference speech from the same speaker clustered together while different speakers are distant apart. Thus, hypernetwork behaves as a dynamic adapter, conditioning its parameter on the speaker style. In contrary to this, a static adapter will learn a single set of parameters (one mark) in the possible space of parameters.
\subsection{Other Discussions}
\paragraph{Layernorms (standard and conditional).} Similar to \citet{mahabadi2021parameter}, we also experiment with inserting a conditional layer norm in the adapter network output where, however, results did not show any significant performance difference both in $\model{}$ and AdapterTTS.
\paragraph{Low-Rank Adaptation.} We refrain from employing hypernetwork for generating LoRA \cite{hulora} parameters instead of the bottleneck adapter. One primary reason is the higher number of parameters involved in LoRA due to the adaptation of query, key, and value in each backbone layer. We leave further exploration in this direction as a future work.

\begin{table*}[h!]
\centering
\small
\resizebox{0.7\textwidth}{!}{
\begin{tabular}{@{}lccccl@{}}
\toprule
\textbf{Model}     & \textbf{COS} $\uparrow$         & \textbf{FFE} $\downarrow$        & \textbf{WER} $\downarrow$ & \textbf{MCD} $\downarrow$ & \textbf{Params} \\ \midrule
\model{}$_e$(2)        & $74.64_{(\pm0.0038)}$ & $37.27_{(\pm0.0178)}$ &  $\textbf{0.2170}$   &  $5.4041$   &   50K (0.14\%)
\\
\model{}$_e$(8)        & $75.43_{(\pm0.0035)}$ & $36.07_{(\pm0.0193)}$ &  $0.2367$   &  $5.3930$   &   151K (0.42\%)
\\
\model{}$_e$(32)       & $\textbf{76.02}_{(\pm0.0035)}$ & $\textbf{35.17}_{(\pm0.0190)}$ &  $0.2449$   &  $\textbf{5.3779}$   &   554K (1.5\%)
\\
\model{}$_e$(128)      & $75.97_{(\pm0.0036)}$ & $35.93_{(\pm0.0193)}$ &  $0.2612$   &  $5.4210$   &   2.17M (6.06\%)
\\ \bottomrule
\end{tabular}
}
\caption{\footnotesize{Varying number of parameters of hypernetwork in encoder with VCTK as the target domain.}}
\label{tab:hypernetwork_ablations_dimension_encoder}
\end{table*}
\begin{table*}[h!]
\centering
\small
\resizebox{0.7\textwidth}{!}{
\begin{tabular}{@{}lccccl@{}}
\toprule
\textbf{Model}     & \textbf{COS} $\uparrow$         & \textbf{FFE} $\downarrow$        & \textbf{WER} $\downarrow$ & \textbf{MCD} $\downarrow$ & \textbf{Params} \\ \midrule
\model{}$_{e/v/d}$(2)        & $77.26_{(\pm0.0034)}$ & $35.10_{(\pm0.0180)}$ & $\textbf{0.2314}$    &  $5.3436$   &   150K (0.42\%) 
\\ 
\model{}$_{e/v/d}$(8)        & $79.46_{(\pm0.0030)}$ & $34.47_{(\pm0.0198)}$ & $0.2340$    &  $5.3293$  &   453K (1.27\%) 
\\
\model{}$_{e/v/d}$(32)       & $80.67_{(\pm0.0027)}$ & $33.85_{(\pm0.0209)}$ & $0.2513$    &  $\textbf{5.2761}$  &   1.66M (4.65\%) 
\\
\model{}$_{e/v/d}$(128)      & $\textbf{81.49}_{(\pm0.0031)}$ & $\textbf{33.58}_{(\pm0.0223)}$ & $0.2622$    &  $5.2879$   &   6.50M (18.19\%) 
\\ \bottomrule
\end{tabular}
}
\caption{\footnotesize{Varying number of parameters of hypernetwork in the encoder, decoder, and VA with VCTK as the target domain.}}
\label{tab:hypernetwork_ablations_dimension_all}
\end{table*}
\section{Conclusion}
In this paper, we present \textbf{\model{}}, an approach that enhances the effectiveness of adapters by conditioning them on speaker embeddings. Utilizing a "hypernetwork" to customize adapter block weights for the TTS backbone network, we significantly expand the adapter parameter space. This dynamic method replaces the conventional static adapter parameter set, enabling input-conditioned parameter sampling. Additionally, the hypernetwork's continuous parameter space theoretically allows the generation of adapter parameters for numerous speakers without increasing hypernetwork parameters. This makes \model{} an excellent choice for multi-speaker TTS adaptation, surpassing traditional adapter limitations.

\textit{Limitations}: While hypernetworks exhibit promising enhancements in both adaptation domains, there are training challenges to address. Time and resource constraints may have led to potential underfitting, negatively impacting performance. Additionally, hypernetworks tend to overfit the backbone model on the adaptation domain, warranting further research to enhance their generalizability. Notably, the relatively higher number of parameters in hypernetworks poses potential inefficiency for low-resource training.
\nocite{*}

\section{Acknowledgments}
We thank the anonymous reviewers for their constructive feedback. This work was supported in part by the National Key Research and Development Program of China under Grant 2022YFF0902701, in part by the National Natural Science Foundation of China under Grants U21A20468, 62372058, U22A201339, in part by the Fundamental Research Funds for the Central Universities under Grant 2020XD-A07-1, in part by BUPT Excellent Ph.D. Students Foundation CX2021229.

This work is also generously supported by the AcRF MoE Tier-2 grant (Project no. T2MOE2008, and Grantor reference no. MOE-T2EP20220-0017) titled: ``CSK-NLP: Leveraging Commonsense Knowledge for NLP'', and the SRG grant id: T1SRIS19149 titled ``An Affective Multimodal Dialogue System''.

\section{Bibliographical References}\label{sec:reference}

\bibliographystyle{lrec-coling2024-natbib}
\bibliography{lrec-coling2024-example}



\end{document}